\documentclass[10pt,twocolumn,letterpaper]{article}

\usepackage{cvpr}              

\usepackage[dvipsnames]{xcolor}

\definecolor{cvprblue}{rgb}{0.21,0.49,0.74}
\usepackage[pagebackref,breaklinks,colorlinks,citecolor=cvprblue]{hyperref}
\usepackage{color, colortbl,booktabs}
\usepackage{wrapfig,lipsum,booktabs}
\usepackage{adjustbox}
\usepackage{multicol}
\usepackage{float}
\usepackage{marvosym}
\usepackage[accsupp]{axessibility} 

\definecolor{cadmiumyellow}{rgb}{1, 0.9216, 0.2314}
\definecolor{Gray}{gray}{0.9}
\usepackage{multirow}
\usepackage{pifont}
\newcommand{\cmark}{\ding{51}}
\newcommand{\xmark}{\ding{55}}
\DeclareFontFamily{U}{matha}{\hyphenchar\font45}
\DeclareFontShape{U}{matha}{m}{n}{%
<-6> matha5 <6-7> matha6 <7-8> matha7
<8-9> matha8 <9-10> matha9
<10-12> matha10 <12-> matha12}{}
\DeclareSymbolFont{matha}{U}{matha}{m}{n}
\DeclareFontSubstitution{U}{matha}{m}{n}
\DeclareMathSymbol{\abxll}{\mathrel}{matha}{"21}
\DeclareMathSymbol{\abxgg}{\mathrel}{matha}{"22}

\makeatletter
\def\blfootnote{\xdef\@thefnmark{}\@footnotetext}
\makeatother

\def\paperID{4323} 
\def\confName{CVPR}
\def\confYear{2024}

\title{Mask Grounding for Referring Image Segmentation}
\author{Yong Xien Chng$^{1,2}$\hspace{0.4cm} Henry Zheng$^{1}$\hspace{0.4cm} Yizeng Han$^{1}$\hspace{0.4cm} Xuchong Qiu$^{2\dagger}$\hspace{0.4cm} Gao Huang$^{1}$\textsuperscript{\Letter}\\
$^1$Department of Automation, BNRist, Tsinghua University\hspace{0.4cm} $^2$Bosch Corporate Research\\
}

\begin{document}
\maketitle
\begin{abstract}
Referring Image Segmentation (RIS) is a challenging task that requires an algorithm to segment objects referred by free-form language expressions. Despite significant progress in recent years, most state-of-the-art (SOTA) methods still suffer from considerable language-image \textit{modality gap} at the pixel and word level. These methods generally 1) rely on sentence-level language features for language-image alignment and 2) lack explicit training supervision for fine-grained visual grounding. Consequently, they exhibit weak object-level correspondence between visual and language features. Without well-grounded features, prior methods struggle to understand complex expressions that require strong reasoning over relationships among multiple objects, especially when dealing with rarely used or ambiguous clauses.
To tackle this challenge, we introduce a novel Mask Grounding auxiliary task that significantly improves visual grounding within language features, by explicitly teaching the model to learn fine-grained correspondence between masked textual tokens and their matching visual objects. Mask Grounding can be directly used on prior RIS methods and consistently
bring improvements. Furthermore, to holistically address the modality gap, we also design a cross-modal alignment loss and an accompanying alignment module. These additions work synergistically with Mask Grounding.
With all these techniques, our comprehensive approach culminates in MagNet (\underline{Ma}sk-\underline{g}rounded \underline{Net}work), an architecture that significantly outperforms prior arts on three key benchmarks (RefCOCO, RefCOCO+ and G-Ref), demonstrating our method's effectiveness in addressing current limitations of RIS algorithms. Our code and pre-trained weights will be released.
\end{abstract}    
\blfootnote{$\dagger$ Project lead.}
\blfootnote{\textsuperscript{\Letter} Corresponding author.}
\section{Introduction}
\label{sec:intro}

\begin{figure}[t]
  \centering
   \includegraphics[width=\linewidth]{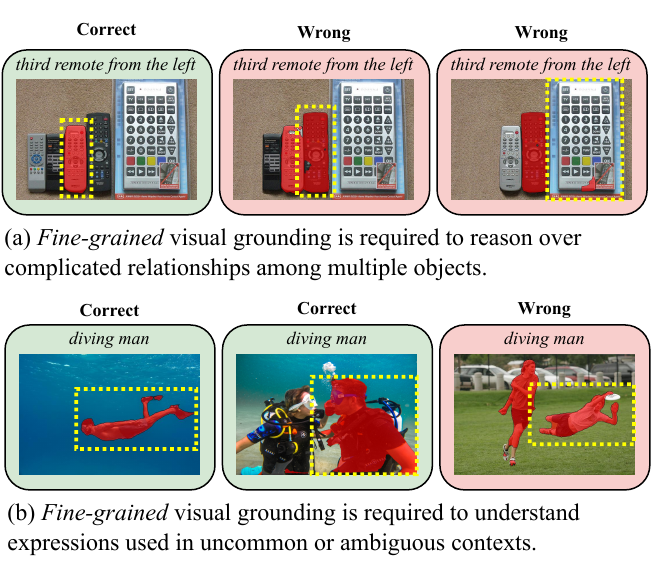}
   \captionsetup{belowskip=-10pt}
   \caption{Importance of Fine-grained Visual Grounding for RIS. Most RIS algorithms lack well-grounded text features. As a result, they struggle in difficult cases illustrated in (a) and (b). \colorbox{red}{Red} mask are predictions of LAVT, one of the recent SOTA RIS methods. \colorbox{cadmiumyellow}{Yellow} dotted boxes are the ground truths.}
   \label{figure1}
\end{figure}

\begin{figure*}
  \centering
  \includegraphics[width=\linewidth]{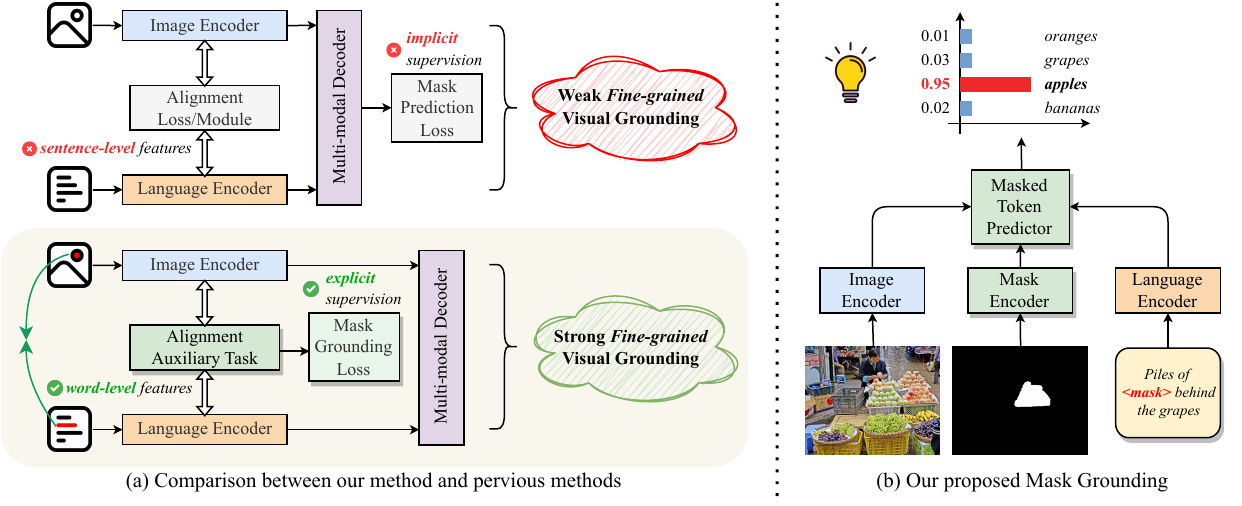}
  \captionsetup{belowskip=-10pt}
  \caption{(a) Current SOTA RIS methods mainly focus on designing and improving multi-modal alignment modules and/or alignment losses. These methods generally 1) do not have explicit training supervision for fine-grained visual grounding and 2) use sentence-level language features or image/pixel-level image features for alignment. As a result, their language features lack precise visual-textual object correspondence. (b) Our proposed Mask Grounding remedies this problem by explicitly teaching our model to learn fine-grained correspondence between masked word tokens and their matching visual objects through an auxiliary alignment task.}
  \label{figure2}
\end{figure*}
Deep learning has greatly improved the performance of vision algorithms on many image segmentation tasks, such as semantic segmentation \cite{fcn,deeplab}, instance segmentation \cite{instance,maskrcnn,dai2016instance,bolya2019yolact} and panoptic segmentation \cite{panoptic,panopticdeeplab}. These tasks require grouping of image pixels under a fixed set of pre-defined categories and mainly differ in the granularity of grouping semantics required. In contrast to these uni-modal segmentation tasks, Referring Image Segmentation (RIS) \cite{ris1,ris2} is a challenging multi-modal task that requires an algorithm to simultaneously understand fine-grained human language expression and make correct pixel-level correspondence to the referred object. Recently, it has gained widespread research attention due to its potential to improve human-robot interaction \cite{robot2}, interactive image editing \cite{edit1,edit2} and advanced driver-assistance systems \cite{dmmi}. 

The key challenge in RIS lies in how to reduce the modality gap between language and image features \cite{lavt,vlt,coupalign}. To tackle this challenge, we need to have an effective alignment between a given language expression and the corresponding image pixels for highlighting the referred target. Ideally, with robust pixel-wise language-image alignment, language and image features should have high feature similarity when referring to the same object and low feature similarity when referring to different objects. However, achieving such alignment is non-trivial because the language expression can be highly complex and diverse. 


As depicted in Fig.~\ref{figure2}, prevailing methods primarily focus on devising innovative losses \cite{coupalign,cris} or introducing new network architectures/modules \cite{lavt,vpd,vlt,mmnet,gres} to bolster language-image alignment. Despite their advancements, two overarching limitations persist. First, these approaches tend to rely on \textit{sentence-level language features} for language-image alignment. Second, they often \textit{lack explicit training supervision} for fine-grained visual grounding. These oversights result in their language features becoming noisy anchors for RIS prediction \cite{ovrcnn,mmnet}, inhibiting the effective learning of fine-grained visual grounding. Consequently, such models face challenges when interpreting referring expressions that require intricate reasoning across complex inter-object relationships or contain clauses used in rare or ambiguous contexts, as exemplified in Fig.~\ref{figure1}.

To address this challenge, we introduce a novel \textbf{Mask Grounding} auxiliary task to explicitly teach our model to make fine-grained correspondence between masked textual tokens and their matching visual objects. Specifically, during training, our model encounters randomly masked textual tokens and has to predict their identities. Instead of relying solely on the surrounding textual context to predict these tokens, our model integrates both visual and segmentation information. This integrated approach is pivotal for the model to make accurate prediction, as it must discern and establish the correct linkage between the masked tokens and their corresponding visual objects. Learning to do so ensures that our model acquires a profound proficiency in the highly-coveted fine-grained visual grounding. The efficacy of Mask Grounding is empirically validated with extensive ablation experiments. Moreover, we also show that Mask Grounding is universal and can be directly used on prior RIS methods to bring significant improvements.

In addition to Mask Grounding, we also design a cross-modal alignment loss and an alignment module to holistically bridge the modality gap. With all these enhancements, our resulting MagNet (\underline{Ma}sk-\underline{g}rounded \underline{Net}work) sets new records by significantly outperforming previous SOTA methods across all key datasets (RefCOCO \cite{refcoco}, RefCOCO+ \cite{refcoco} and G-Ref \cite{gref,umd}). Notably, our method consistently outperforms these SOTA methods by large margins of up to \textbf{2.48} points in overall IoU. Visual examination of MagNet's predictions reinforces our claim and shows that our method works well in complex scenarios. 


Our main contributions are summarized as follows:
\begin{enumerate}
    \item We highlight the shortcomings in recent state-of-the-art (SOTA) RIS algorithms, pinpointing the lack of fine-grained visual grounding. 
    \item We introduce the Mask Grounding auxiliary task, a novel method aimed at enhancing fine-grained visual grounding in existing RIS algorithms. Its effectiveness is validated through rigorous ablation studies.
    \item Using Mask Grounding, together with our specially designed cross-modal alignment loss and an accompanying alignment module, we present MagNet (\underline{M}ask-\underline{g}rounded \underline{Net}work), a new SOTA network for RIS.
\end{enumerate}


\section{Related works}
\label{sec:formatting}

\textbf{Architecture Design for RIS.} Early works \cite{ris1,rmi,rrn,kwa} follow a concatenate-then-convolve pipeline, where language and image features are fused by concatenation. Subsequent works \cite{rmi,rrn,step,sadlr} improve upon this pipeline by using RNN or dynamic networks~\cite{han2021dynamic,han2021spatially,han2022latency,han2023latency} to progressively refine the segmentation mask. Other works \cite{lavt,efn} investigate the position to perform language-image fusion and conclude that early fusion performs the best. Apart from designing novel fusion mechanisms, some works \cite{cmpc,lscm,mattnet} exploit known linguistic structures or object relationships to enhance language-image fusion. Riding on the success of attention architecture \cite{transformer, vit,han2023dynamic}, current works mostly use unidirectional \cite{kwa, lavt, cmsa} or bidirectional \cite{brinet, coupalign} cross-attention modules to perform language-image fusion. To improve model performance on novel composition of learned concepts, a recent work \cite{mcres} uses meta learning \cite{maml}. Driven by the success of large language models \cite{gpt,llama}, newer works \cite{pvd, polyformer,seqtr} explore formulating RIS as an auto-regressive vertex generation problem. Lately, VPD \cite{vpd} attempts to exploit semantic information in diffusion models \cite{ddpm,sde,Ni2024Revisit} for RIS, whereas ReLA \cite{gres} and DMMI \cite{dmmi} generalize RIS to support an arbitrary number of targets. Despite huge progress in RIS architecture design, prior studies often expect language-image alignment to be performed implicitly through mask prediction. We enhance this by introducing an auxiliary task for explicit language-image feature alignment.

\noindent \textbf{Loss Design for RIS.} Early works train RIS models with simple binary cross entropy loss. Inspired by the success of prior works \cite{wang2021exploring,zhao2021contrastive} in adopting contrastive loss \cite{cl,clip,moco,simclr} for semantic segmentation tasks, recent works \cite{coupalign, vlt} start to use contrastive loss in order to regularize the segmentation embedding space and achieve good results. Contrary to prior works that use global-pooled language features for loss computation, we focus on learning fine-grained object correspondence at the pixel-word level.

\noindent \textbf{Masked Language Modeling.} \underline{M}asked \underline{l}anguage \underline{m}odeling (MLM) is a powerful technique for natural language processing that trains a model to restore missing or corrupted tokens in an input text. It was introduced by BERT \cite{bert} and has become a popular technique for pre-training language \cite{flan,opt} and visual language \cite{blip2,vlbert,vilt} models. Recently, it has been shown to scale excellently \cite{align, gpt} and generalize well to various downstream tasks \cite{gpt,llama}. A work closely related to ours is MaskedVLM \cite{mvlm}, which is a multi-modal adaptation of MLM that jointly performs masked vision and language modeling. It does so by reconstructing the masked signal of one modality with the help from the another modality. Mask Grounding differs from MaskedVLM by using extra mask signals that directly match the missing words to ensure clear and meaningful reconstructions, so that fine-grained visual grounding can be effectively learnt.


\section{Method}

In this section, we first describe our architecture overview (Sec.~\ref{section3.1}). Then, we explain our proposed Mask Grounding (Sec.~\ref{section3.2}) auxiliary task, cross-modal alignment module (Sec.~\ref{section3.3}) and alignment loss (Sec.~\ref{section3.4}). Finally, we give the overall loss function Sec.~\ref{section3.5} for our model.
\subsection{Architecture Overview}
\label{section3.1}





MagNet (\underline{Ma}sk-\underline{g}rounded \underline{Net}work) adopts a unified approach that integrates three inter-linked modules to enhance visual-textual object correspondence and segmentation accuracy. Mask Grounding is the first of these integrated modules, designed to improve fine-grained visual grounding in language features. It accomplishes this by teaching the model to predict masked textual tokens, using a combination of visual cues, linguistic context, and segmentation information. Building upon Mask Grounding's enriched language features, Cross-modal Alignment Module (CAM) steps in to fine-tune the bi-directional interaction between the refined language and image features. By incorporating global contextual information from multiple image scales, CAM ensures that the multi-modal features are in sync, addressing the granularity discrepancies between textual descriptions and visual information. Finally, Cross-modal Alignment Loss (CAL) cohesively weaves together pixel-to-pixel and pixel-to-text alignments. By simultaneously considering these alignments, CAL ensures that segments created by the model are not only accurate in shape but also correctly match their referring textual descriptions.

\begin{figure*}
  \centering
  \includegraphics[width=\linewidth]{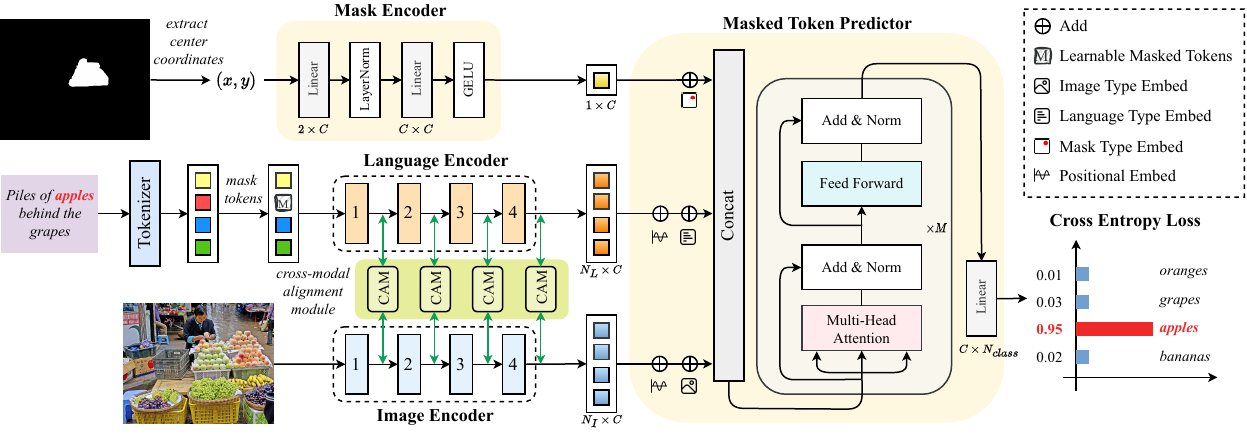}
    \caption{Overview of Mask Grounding. This task enriches fine-grained visual grounding in language features by guiding the model to learn detailed textual-visual associations. To perform this task, we first use an MLP-based Mask Encoder to encode center-coordinates of segmentation masks. Then, we randomly mask textual tokens in language inputs before extracting their features. Finally, we pass the encoded language, image and mask features to a Transformer-based Masked Token Predictor to perform masked token prediction.}
  \label{figure3}
    \vskip -0.1 in
\end{figure*}

\subsection{Mask Grounding} 
\label{section3.2}
Inspired by prior works \cite{mae,bert,beit3} that have shown the effectiveness of using masked input modeling to learn good feature representation, we propose a novel Mask Grounding auxiliary task to improve the learning of fine-grained visual grounding information in language features. As shown in Fig.~\ref{figure3}, given an input image, its corresponding referring expression and segmentation mask, we randomly replace some tokens in the tokenized referring expression with a special learnable mask token and train our model to predict the actual tokens being masked. By successfully predicting the identities of masked tokens, our model will acquire the ability to understand which parts of the text correspond to which parts of the image, thus learning fine-grained visual grounding in the process. Specifically, to perform this auxiliary task, we first encode the segmentation mask into a mask embedding by first extracting the center coordinates of the mask region and passing it through a 2-layer MLP. At the same time, we use a linear layer to project the language embedding into the same dimension as the image embedding. Then, we employ the proposed Masked Token Predictor to jointly process all these concatenated embeddings with attention mechanism for masked token prediction. Finally, a cross-entropy loss $\mathcal{L}_\text{grounding}$ is used to compare the final predicted distribution with the target distribution. The large-scale BERT \cite{bert} vocabulary is adopted as our word class list, as it is generally accepted to have open-vocabulary capability. Although additional forward pass through the language encoder is required to process the masked expression, overall computational cost only increase by 4.76\% as the language encoder is very small. We believe this slight increase in computational cost is an acceptable trade-off to improve visual grounding in language features. 

A brief mathematical formulation for Mask Grounding can be given as follows: Let \textbf{T}, $\textbf{I}$, \textbf{M} be the input to the language encoder, image encoder and mask encoder,
\begin{gather}
\textbf{O} = \text{LanguageEncoder}(\text{Mask}(\textbf{T})), \\
\textbf{P}= \text{ImageEncoder}(\textbf{I}), \\
\textbf{C}=\text{MaskEncoder}(\textbf{M}), \\
\mathcal{L}_{\text{grounding}} = \mathcal{L}_\text{CE}(\textbf{y}_{\text{gt}}, \text{Predictor}(\text{Concat}([\textbf{O}, \textbf{P}, \textbf{C}]),
\end{gather}
where Predictor is a BERT \cite{bert}-like encoder, M is the center coordinates of ground truth masks, $\textbf{y}_{\text{gt}}$ is the label of the masked token and $\mathcal{L}_\text{CE}$ is the cross entropy loss. In our experiments, we use Swin-B \cite{swin} as our image encoder, and BERT-base \cite{bert} as our language encoder, but our approach is not specifically bound to these encoders.

\noindent \textbf{Discussion.} In Tab.~\ref{table3}(a), we demonstrate Mask Grounding's superiority over both the standard masked language modeling (MLM) \cite{bert,llama,gpt,align} and masked-vision language modeling (MaskedVLM) \cite{mvlm}, highlighting our approach's effectiveness. Our advantages over these techniques include: 1) \textit{Modality Integration}: Traditional MLM is uni-modal and lacks correspondence between referring expressions and their matching visual objects.. While MaskedVLM is multi-modal, Mask Grounding surpasses it by introducing an additional masking signal that aligns with the masked words and their matching visual objects, enabling a more coherent reconstruction. This approach exposes word-object correspondence and allows fine-grained visual grounding to be learnt. 2) \textit{Task Nature}: MLM and MaskedVLM serve as general pre-training tasks and require fine-tuning for specific downstream applications. In contrast, Mask Grounding is designed as a specialized auxiliary task for RIS, enhancing fine-grained visual grounding within language features right from the training phase. Consequently, there is no need for additional fine-tuning. 3) \textit{Prediction Context}: While MLM and MaskedVLM predict using textual or textual-visual contexts, Mask Grounding incorporates both with additional segmentation information. By leveraging this additional information, our model can outperform prior methods in complex scenarios where text and visual elements are closely intertwined. For instance, consider the scenario illustrated in Fig.~\ref{figure3}. When the term ``apples" in ``piles of apples behind the grapes" is masked, a model lacking precise word-object correlation might falter in predicting the appropriate term. Several other words might yield a semantically consistent sentence, but they would not be accurate in the given visual context.

\subsection{Cross-modal Alignment Module}
\label{section3.3}
\begin{figure}[h]
  \centering
  \includegraphics[width=\linewidth]{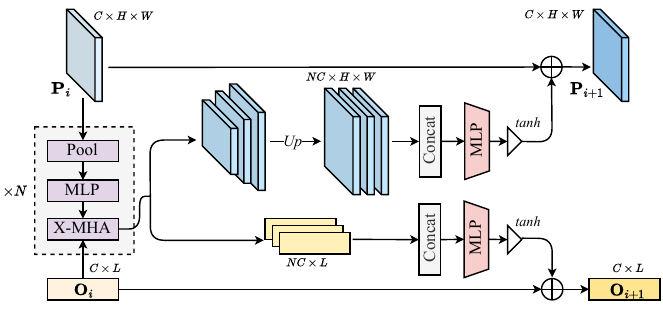}
  \caption{Cross-modal Alignment Module. This module enables bidirectional language-image interaction and addresses granularity mismatches between language and image features, thereby enhancing segmentation accuracy for RIS. $\text{X-MHA}$ denotes bi-directional cross-modal multi-head attention. $\textbf{P}_i$ and $\textbf{P}_{i+1}$ denote input and output image features, whereas $\textbf{O}_i$ and $\textbf{O}_{i+1}$ denote input and output language features. Up denotes upsampling.}
  \label{figure4}
  \vskip -0.1 in
\end{figure}

To further improve the performance of our model, we also make a meaningful improvement to the popular cross-modal alignment mechanism proposed by prior work \cite{lavt}. As depicted in Fig.~\ref{figure4}, our \underline{c}ross-modal \underline{a}lignment \underline{m}odule (CAM) improves language-image alignment by injecting global contextual prior into image features before performing language-image fusion. CAM first uses pooling operations with different window sizes to generate $K$ feature maps of different pyramid scales. Then, each of these feature maps passes through a 3-layer MLP to better extract global information, before cross-attending with the opposite modality. After that, all the output features are upsampled to the original feature map via bilinear interpolation and concatenated along the channel dimension. A 2-layer MLP is subsequently used to to reduce the channel dimension of this concatenated feature back to the original one. To prevent the multi-modal signal from overwhelming the original signal, a gate with Tanh nonlinearity is used to modulate the final output. Finally, this post-gate feature is added back to the input feature before being passed to the next stage of the image or language encoder. We split language encoder into 4 stages with an equal number of layers and add CAM to the end of every stage of image and language encoder. 

Mathematically, CAM can be represented as follows: Let $\textbf{T}_i$ and $\textbf{I}_i$ be the text/image input to each stage of the language and image encoder. At each stage,
\begin{gather}
\textbf{O}_i = \text{LanguageStage}(\textbf{T}_i),\, \textbf{P}_i = \text{ImageStage}(\textbf{I}_i), \\
\textbf{P}_i^k = \text{MLP}_k(\text{Pool}_k(\textbf{P}_i)), \\
\textbf{O}^K_{i,p2t}, \textbf{P}^k_{i,t2p} = \text{X-MHA}_k(\textbf{O}_i, \textbf{P}^k_n), \\
\textbf{O}_{i,p2t} = \text{Concat}([\textbf{O}^i_{i,p2t},...,\textbf{O}^N_{i,p2t}], \\
\textbf{P}_{i,t2p} = \text{Concat}([\text{Up}(\textbf{P}^1_{i,t2p},...,\text{Up}(\textbf{P}^N_{i,t2p})], \\
\textbf{O}_{i+1} = \textbf{O}_i + \text{tanh}(\text{MLP}(\textbf{O}_{i,p2t})), \\
\textbf{P}_{i+1} = \textbf{P}_i + \text{tanh}(\text{MLP}(\textbf{P}_{i,t2p})),
\end{gather}
\noindent where Up denotes upsampling and X-MHA \cite{glip} denotes bi-directional cross-modal multi-head attention.



CAM enhances cross-modal alignment by enabling bi-directional language-image interaction, which stands in contrast to the widely-used one-way language to image alignment module proposed by LAVT \cite{lavt}. Moreover, CAM adopts a pyramid pooling technique to utilize multi-scale average-pooled image features. This technique adeptly resolves the granularity mismatch issue by capturing image features at multiple scales, allowing our network to handle the varied levels of detail present in language descriptions. This is particularly beneficial for RIS, where the model must accurately interpret and segment according to a diverse range of descriptive queries.


\subsection{Cross-modal Alignment Loss}
\label{section3.4}

On top of that, similar to previous works \cite{cris, coupalign}, we also use cross-modal alignment loss to explicitly align language and image features. Our \underline{c}ross-modal \underline{a}lignment \underline{l}oss (CAL) is holistic and consider both pixel-to-pixel ($\mathcal{L}_{\text{P2P}}$) and pixel-to-text ($\mathcal{L}_{\text{P2T}}$) consistency. 

Mathematically, CAL is computed as follows: Given language feature $\textbf{T} \in \mathbb{R}^{M \times D}$ produced by the language encoder and final pixel decoder mask feature $\textbf{I} \in \mathbb{R}^{C_L \times H_L \times W_L}$ with $|\mathcal{P}|$ positive pixel features, $|\mathcal{N}|$ negative pixel features, let $\textbf{I}^+_i$ be the $i^{th}$ pixel feature in the positive set $\mathcal{P}$, $\textbf{I}^-_j$ be the $j^{th}$ pixel feature in the background set $\mathcal{N}$ and $\textbf{T}_k$ be the $k^{th}$ language token, then
\begin{gather}
\mathcal{L_{\text{CAL}}} = \mathcal{L}_{\text{P2P}} + \mathcal{L}_{\text{P2T}} \\
\begin{aligned}
\mathcal{L}_{\text{P2P}} &= -\frac{1}{|\mathcal{P}|} \sum_i^{|\mathcal{P}|} \frac{e^{\textbf{I}^+_i \cdot \textbf{I}_\text{avg}^+/\tau_1}}{e^{\textbf{I}^+_i \cdot \textbf{I}_\text{avg}^+/\tau_1} + \sum_j^{|\mathcal{N}|} e^{\textbf{I}^+_i \cdot \textbf{I}^-_j / \tau_1}} \\
&+ -\frac{1}{|\mathcal{N}|} \sum_j^{|\mathcal{N}|} \frac{e^{\textbf{I}^-_j \cdot \textbf{I}_\text{avg}^- / \tau_1}}{e^{\textbf{I}^-_j \cdot \textbf{I}_\text{avg}^-/\tau_1} + \sum_i^{|\mathcal{P}|} e^{\textbf{I}^-_j \cdot \textbf{I}^+_i/\tau_1}},
\end{aligned} \\
\begin{aligned}
\mathcal{L}_{\text{P2T}} &= -\frac{1}{|\mathcal{P}|} \sum_i^{|\mathcal{P}|} \frac{e^{\textbf{I}^+_i \cdot \textbf{T}_\text{avg}/\tau_2}}{e^{\textbf{I}^+_i \cdot \textbf{T}_\text{avg}/\tau_2} + \sum_j^{|\mathcal{N}|} e^{\textbf{I}^+_i \cdot \textbf{I}^-_j/\tau_2}},
\end{aligned} 
\end{gather}
where $\textbf{I}_\text{avg}^+ = \frac{1}{|\mathcal{P}|} \sum_i^{|\mathcal{P}|} \textbf{I}^+_i$ and $\textbf{I}_\text{avg}^- = \frac{1}{|\mathcal{N}|} \sum_j^{|\mathcal{N}|} \textbf{I}^-_j$ are the average pooled positive and negative pixel features, $\textbf{T}_\text{avg} = \text{proj}(\frac{1}{M} \sum_m^{M} \textbf{T}_k)$ is the average pooled and linearly projected word feature and $\tau_1$, $\tau_2$ are hyper-parameters that affect the sharpness of the probability distribution. Note that all language and image features are L2-normalized before any dot product computation, but not explicitly shown in the equations above for brevity.

CAL differs from alignment losses used in prior works \cite{cris, coupalign} by holistically integrating both pixel-to-pixel and pixel-to-text alignments within a single cohesive system. Precise pixel-to-pixel alignment ensures that segmentation outputs have accurate shapes and boundaries, whereas precise pixel-to-text alignment enables our model to correctly associate textual descriptions with their matching image regions. This dual alignment mechanism allows our model to effectively parse and interpret the nuanced interplay between image details and language cues, leading to more accurate and contextually relevant segmentation outputs.

\begin{table*}
    \centering

    \resizebox{0.96\linewidth}{!}{
    \begin{tabular}{c|c|c|ccc|ccc|ccc}
        \toprule
        &\multirow{2}{*}{Method}& \multirow{2}{*}{Backbone}  & \multicolumn{3}{c|}{RefCOCO (easy)} & \multicolumn{3}{c|}{RefCOCO+ (medium) } & \multicolumn{3}{c}{G-Ref (hard)} \\
        &&& val & test A & test B & val & test A & test B & val (U) & test (U) & val (G)\\
        \midrule

        & VLT \cite{vlt}  & Darknet-53  
                & 65.65 & 68.29 & 62.73
                & 55.50 & 59.20 & 49.36
                & 52.99 & 56.65 & 49.76\\

         & ReSTR \cite{restr} & ViT-B-16 
                & 67.22 & 69.30 & 64.45
                & 55.78 & 60.44 & 48.27
                &- & - & 54.48\\

         & CRIS \cite{cris} & ResNet-101 
                & 70.47 & 73.18 & 66.10
                & 62.27 & 68.08 & 53.68
                & 59.87 & 60.36 & -\\

          & LAVT \cite{lavt} & Swin-B 
                & 72.73 & 75.82 & 68.79
                & 62.14 & 68.38 & 55.10
                & 61.24 & 62.09 & 60.50\\
         Single & VPD \cite{vpd} & Swin-B 
                &73.46 & - & - 
                &63.93 & - & -
                &63.12 &- &-\\
         Dataset & CoupAlign \cite{coupalign} & Swin-B 
                & 74.70 & 77.76 & 70.58
                & 62.92 & 68.34 & 56.69
                & 62.84 & 62.22 & - \\
         & PVD \cite{pvd} & Swin-B 
                & 74.82 & 77.11 & 69.52
                & 63.38 & 68.60 & 56.92
                & 63.13 & 63.62 & 61.33\\
        & SADLR \cite{sadlr} & Swin-B 
                & 74.24 & 76.25 & 70.06
                & 64.28 & 69.09 & 55.19
                & 63.60 & 63.56 & 61.16\\
        & MCRES \cite{mcres} & Swin-B 
                & 74.92 & 76.98 & 70.84
                & 64.32 & 69.68 & 56.64
                & 63.51 & 64.90 & 61.63\\
        & ReLA \cite{gres} & Swin-B 
                & 73.82 & 76.48 & 70.18
                & 66.04 & 71.02 & 57.65
                & 65.00 & 65.97 & 62.70\\
        \cmidrule{2-12}

        \rowcolor{Gray} \cellcolor{White} &  MagNet (Ours) &  Swin-B 
                &  \textbf{75.24} &  \textbf{78.24} &  \textbf{71.05}
                &  \textbf{66.16} &  \textbf{71.32} &  \textbf{58.14}
                &  \textbf{65.36} &  \textbf{66.03} &  \textbf{63.13}\\
        \midrule
        & SEEM$^\dagger$  \cite{seem} & Focal-T  
                & - & - & -
                & - & - & -
                & 65.7 & - & - \\
         Multiple / Extra & LISA-7B$^\dagger$ \cite{lisa} & SAM-H 
                & 74.1 & 76.5 & 71.1
                & 62.4 & 67.4 & 56.5
                & 66.4 & 68.5 & - \\
         Datasets & PolyFormer$^\dagger$ \cite{polyformer} & Swin-B   
                & 74.82 & 76.64 & 71.06
                & 67.64 & 72.89 & 59.33
                & 67.76 & 69.05 & - \\
        \cmidrule{2-12}
        
        \rowcolor{Gray} \cellcolor{White} & MagNet$^\ddagger$ (Ours) &  Swin-B 
                & \textbf{76.55} & \textbf{78.27} &  \textbf{72.15}
                & \textbf{68.10} &  \textbf{73.64} & \textbf{61.81}
                &  \textbf{67.79} &  \textbf{69.29} &  -\\ 
        \bottomrule
    \end{tabular}}
    \caption{Comparison with SOTA methods using the oIoU metric. \textit{Single dataset} refers to strictly following the predefined train/test splits of the original RefCOCO, RefCOCO+ and G-Ref datasets. \textit{Multiple datasets} refers to combining the train splits from these 3 datasets with test images removed to prevent data leakage. \textit{Extra datasets} refers to using additional data beyond RefCOCO, RefCOCO+ and G-Ref. $^\dagger$ indicates models that use extra datasets. $^\ddagger$ indicates that our model only uses multiple datasets. \textbf{Bold} indicates best.}
    \label{table1}
     \vskip -0.15 in
\end{table*}

\subsection{Loss Function}
\label{section3.5}
Our loss function is a weighted combination of the following 4 different losses:
\begin{equation}
\begin{gathered}
\mathcal{L} = \lambda_{\text{BCE}} \mathcal{L}_{\text{BCE}} + \lambda_{\text{Dice}} \mathcal{L}_{\text{Dice}} + \\ \lambda_{\text{CAL}} \mathcal{L}_{\text{CAL}} + \lambda_{\text{grounding}} \mathcal{L}_{\text{grounding}},
\end{gathered}
\end{equation}
with  $\lambda_{\text{BCE}} = 2.0$, $\lambda_{\text{Dice}} = 2.0$, $\lambda_{\text{CAL}} = 0.5$, and $\lambda_{\text{grounding}} = 1.0$ for all our experiments.

\section{Experiments}

In this section, we first describe the datasets and evaluation metrics (Sec.~\ref{section4.1}). Then, we compare our method with SOTA RIS methods (Sec.~\ref{section4.4}). Finally, we show some visualization results (Sec.~\ref{section4.5}) and ablate our proposed method (Sec.~\ref{section4.6}). Due to space limitation, exact implementation details of our method are relegated to the Supplementary.

\subsection{Datasets and Evaluation Metrics}
\label{section4.1}
We evaluate our proposed method on three standard benchmark datasets: RefCOCO \cite{refcoco}, RefCOCO+ \cite{refcoco}, and G-Ref \cite{gref,umd} using three commonly used metrics: overall intersection-over-union (oIoU), mean intersection-over-union (mIoU), and precision values at 0.5, 0.7, and 0.9 IoU threshold levels (P@X). More details regarding these datasets and metrics can be found in the Supplementary.




\subsection{Main Results}
\label{section4.4}

\begin{figure*}
\centering
\includegraphics[width=0.85\textwidth]{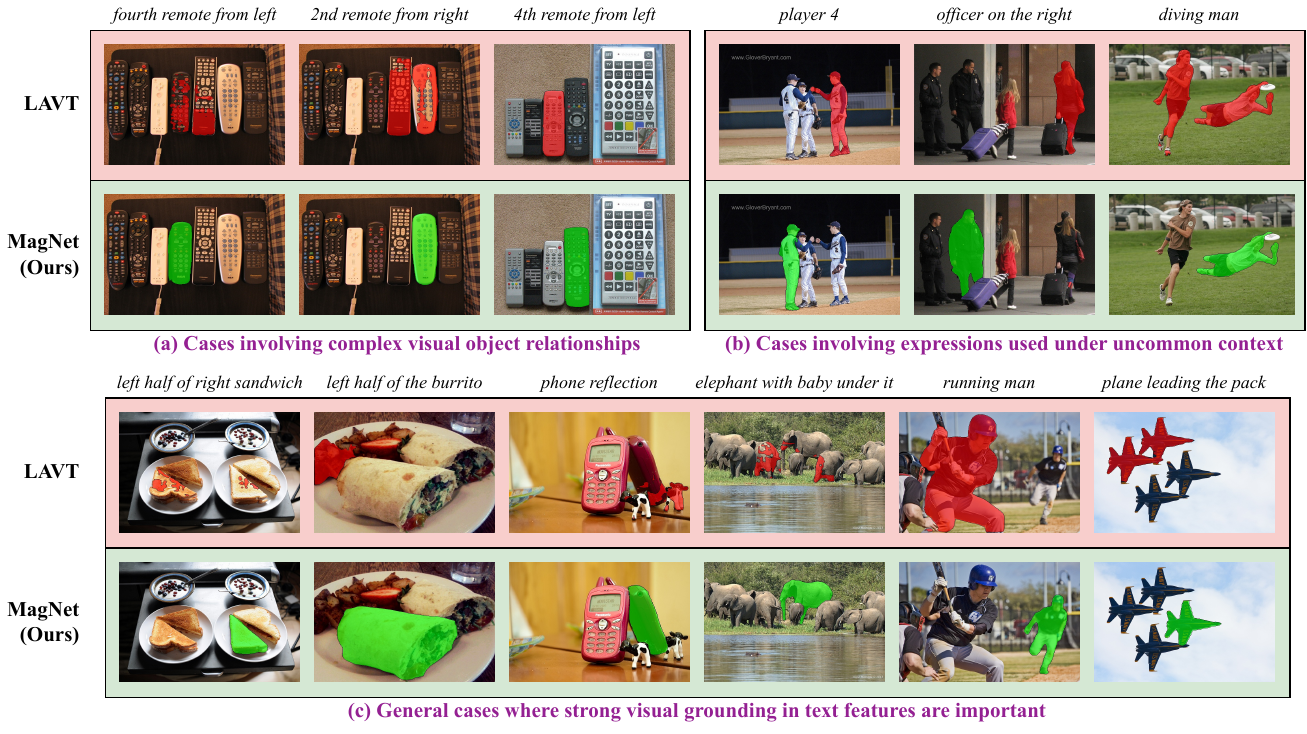}
\captionsetup{belowskip=-10pt}
\caption{Visualization of MagNet's predictions. Compared to one of the state-of-the-art method, LAVT, our method performs much better in various complex scenerios, suggesting its impressive capability to reason about various complex visual-object relationships.}
\label{visualization}
\end{figure*}

In Tab.~\ref{table1}, we evaluate MagNet against other SOTA methods on RefCOCO \cite{refcoco}, RefCOCO+,\cite{refcoco} and G-Ref \cite{gref,umd} datasets using the oIoU metric. Under the \textit{single dataset} setting, MagNet is the first method that consistently outperforms all previous methods on all evaluation subsets of these datasets. Previous methods usually overfit to one of these benchmarks and perform worse in others. Remarkably, on RefCOCO, MagNet outperforms the very recent SOTA RIS method ReLA \cite{gres} by considerable margins of \textbf{1.42}, \textbf{1.76}, and \textbf{0.87} points on the validation, testA, and testB subsets, respectively. To have a more comprehensive evaluation of our method, we also assess MagNet using other metrics and display the results on Tab.~\ref{table2}. As shown, MagNet has much better mIoU and P@X performance than all previous SOTA methods. In particular, our method surpasses previous SOTA methods by \textbf{0.32} points on the oIoU metric, \textbf{0.91} points on the mIoU metric and \textbf{0.89}, \textbf{1.25}, \textbf{1.41} points on the precision metric at 0.5, 0.7 and 0.9 IoU threshold levels. Under the \textit{multiple / extra datasets} setting, our method also surpasses recent SOTA methods \cite{polyformer,lisa} that use large language models \cite{llama} or has much slower inference speed, by large margins of up to \textbf{2.48} points.  

\begin{table}[h]
\centering
    \resizebox{\linewidth}{!}{
    \begin{tabular}{c|ccccc}
        \toprule
        \multirow{2}{*}{Method}& \multicolumn{5}{c}{RefCOCO val} \\
        & oIoU & mIoU & P@0.5  & P@0.7 & P@0.9  \\
        \midrule
        LAVT \cite{lavt} 
                & 72.73 & 74.46 & 84.46 	
                & 75.28 & 34.30  \\
        ReLA \cite{gres} 
                & 73.82 & 75.61 & 85.82
                & 77.71 & 34.99   \\
        CoupAlign \cite{coupalign} 
                & 74.70 & 75.49 & 86.40
                & 77.59 & 32.40  \\
        MCRES \cite{mcres} 
                & \underline{74.92} & - & 86.23
                & 77.25 & 35.61   \\
        SADLR \cite{sadlr} 
                & 74.24 & \underline{76.52} & \underline{86.90}
                & \underline{78.76} & \underline{37.36}   \\

        \midrule
        \rowcolor{Gray}
        MagNet (ours) 
                & \textbf{75.24} & \textbf{77.43} & \textbf{87.79} & \textbf{80.01} & \textbf{38.77}  \\
        \bottomrule
    \end{tabular}
    }
    
    \caption{Comparison with SOTA methods on RefCOCO val using oIoU, mIoU and P@X (Precision at IoU threshold value X). \textbf{Bold} indicates best and \underline{underline} indicates second best.}
    \label{table2}
     \vskip -0.15 in
  \end{table}  
  
\subsection{Visualizations}

\label{section4.5}
In Fig. \ref{visualization}, we show some representative mask predictions of MagNet and LAVT \cite{lavt} on RefCOCO validation set. Here, we only compare with LAVT because it is a SOTA method that provides reproducible codes and pre-trained weights. MagNet outperforms LAVT in scenes that involve complex visual-object relationships, contain uncommon expressions or require strong visual grounding information. Impressively, MagNet demonstrates ability to grasp complex visual cues such as \textit{reflection}, \textit{leading} and \textit{running}.

\subsection{Ablation Studies}
\label{section4.6}
In this section, we investigate the effectiveness of all core components of our model. For experimental efficiency, we use a shorter training schedule of 10 epochs and smaller input images of $224 \times 224$, causing the results to be different from Tab.~\ref{table1}. Other experimental settings are kept the same. We reproduce the numbers for LAVT \cite{lavt}, ReLA \cite{gres} and CRIS \cite{cris} using their official codes. All ablations are performed on validation splits of RefCOCO and RefCOCO+.

\begin{table*}
    \begin{subtable}[t]{0.3\textwidth}
        \centering
        \resizebox{\linewidth}{!}{
        \begin{tabular}{l | cc}
    \toprule
\multicolumn{1}{c|}{Model} & RefCOCO & RefCOCO+ \\
\midrule
Baseline   &67.08 & 55.98\\
Baseline + MLM & 67.31 \textcolor{ForestGreen}{(+0.23)} & 55.69 \textcolor{BrickRed}{(-0.29)} \\
Baseline + MaskedVLM & 67.33 \textcolor{ForestGreen}{(+0.25)} & 56.13 \textcolor{ForestGreen}{(+0.15)} \\
\rowcolor{Gray} Baseline + MG &  \textbf{68.52} \textcolor{ForestGreen}{(\textbf{+1.44})} & \textbf{57.26} \textcolor{ForestGreen} {(\textbf{+1.28})}\\
\bottomrule
\end{tabular}}
\captionsetup{justification=centering}
\caption{Effectiveness of Mask Grounding over masked language modeling (MLM) and masked vision-language modeling (MaskedVLM).} 
\end{subtable}
\hfill
        \begin{subtable}[t]{0.3\textwidth}
        \centering
        
        \resizebox{\linewidth}{!}{\begin{tabular}{c | cc}
        \toprule
 Type & RefCOCO & RefCOCO+ \\
\midrule
None     & 67.08 & 55.98 \\
Average &   68.19 \textcolor{ForestGreen}{(+1.11)} & 56.98 \textcolor{ForestGreen}{(+1.00)} \\
\rowcolor{Gray}  Center  & \textbf{68.52}  \textcolor{ForestGreen}{(\textbf{+1.44})} & \textbf{57.26} \textcolor{ForestGreen} {(\textbf{+1.28})}  \\
\bottomrule
\end{tabular}}
       
\caption{Comparing different mask encoder input in Mask Grounding. \textit{Center} denotes center coordinates of masked region. \textit{Average} denotes average visual features within masked region.}
       
    \end{subtable}
    \hfill
    \begin{subtable}[t]{0.3\textwidth}
        \centering
        \resizebox{\linewidth}{!}{\begin{tabular}{c | ccc}
        \toprule
MLP layers & RefCOCO & RefCOCO+ \\
\midrule
None & 67.08 & 55.98 \\
4  & 67.14 \textcolor{ForestGreen}{(+0.06)} & 57.19 \textcolor{ForestGreen}{(+1.21)} \\
\rowcolor{Gray} 8 & \textbf{68.52} \textcolor{ForestGreen}{(\textbf{+1.44})} & \textbf{57.26} \textcolor{ForestGreen} {(\textbf{+1.28})} \\
12 & 66.92 \textcolor{BrickRed}{(-0.16)} &  55.95 \textcolor{BrickRed}{(-0.03)} \\
\bottomrule
\end{tabular}}


    \captionsetup{justification=centering}
       \caption{Sensitivity of Mask Grounding's masked token predictor to different MLP layers.}
    \end{subtable}
    \hfill
    \begin{subtable}[t]{0.3\textwidth}
        \centering
        \resizebox{\linewidth}{!}{
        \begin{tabular}{c | c|c}
    \toprule
 Model & RefCOCO & RefCOCO+\\
\midrule
LAVT \cite{lavt}  & 67.08 & 55.98\\
\rowcolor{Gray} LAVT + MG  &  \textbf{68.52} \textcolor{ForestGreen}{(\textbf{+1.44})} & \textbf{57.26}  \textcolor{ForestGreen}{(\textbf{+1.28})}\\
\midrule
ReLA \cite{gres} & 66.23 & 53.69 \\
ReLA + MG & \textbf{67.33} \textcolor{ForestGreen}{(\textbf{+1.10})} & \textbf{55.21} \textcolor{ForestGreen} {(\textbf{+1.52})} \\
\midrule
CRIS \cite{cris} & 64.67 & 54.84\\
CRIS + MG  & \textbf{65.56} \textcolor{ForestGreen}{(\textbf{+0.89})} &  \textbf{56.23} \textcolor{ForestGreen}{(\textbf{+1.39})}\\
\bottomrule
\end{tabular}}
\caption{Universality of Mask Grounding.} 
\end{subtable}
    \hfill
\begin{subtable}[t]{0.31\textwidth}
        \centering
        \resizebox{\linewidth}{!}{
        \begin{tabular}{c | cc}
\toprule
Pyramid scales & RefCOCO & RefCOCO+ \\
\midrule
None & 67.08& 55.98 \\
\{1\} &67.18 \textcolor{ForestGreen}{(+0.10)}& 56.20 \textcolor{ForestGreen}{(+0.22)}\\
\{1,2\}& 67.44 \textcolor{ForestGreen}{(+0.36)} & 56.46 \textcolor{ForestGreen}{(+0.48)}\\
\{1,2,3\} &  67.79 \textcolor{ForestGreen}{(+0.71)}& 56.74 \textcolor{ForestGreen}{(+0.76)}  \\
\rowcolor{Gray}
\{1,2,3,6\} & \textbf{68.06} \textcolor{ForestGreen}{(+\textbf{0.98})}  &  \textbf{57.04} \textcolor{ForestGreen}{ (+\textbf{1.06})} \\
\bottomrule
\end{tabular}}
\caption{Effectiveness of language-image Cross-modal Alignment Module at different scales.}  
\end{subtable}
\hfill
    \begin{subtable}[t]{0.31\textwidth}
        \centering
        \resizebox{\linewidth}{!}{\begin{tabular}{cc | cc}
        \toprule
$\mathcal{L}_\text{P2P}$ & $\mathcal{L}_\text{P2T}$ & RefCOCO & RefCOCO+ \\
\midrule
\xmark & \xmark  & 67.08 & 55.98\\
\cmark & \xmark  & 67.63 \textcolor{ForestGreen}{(+0.55)} & 56.87  \textcolor{ForestGreen}{(+0.89)} \\
\xmark  & \cmark & 68.27 \textcolor{ForestGreen}{(+1.19)}&  57.22 \textcolor{ForestGreen}{(+1.24)} \\
\rowcolor{Gray}
\cmark  & \cmark  & \textbf{68.44} \textcolor{ForestGreen}{(\textbf{+1.36})} & \textbf{57.61} \textcolor{ForestGreen}{(\textbf{+1.63})} \\
\bottomrule
\end{tabular}} 
\captionsetup{justification=centering}
\caption{Effectiveness of language-image Cross-modal Alignment Loss.}
    \end{subtable}

\captionsetup{belowskip=-10pt}    
\caption{Ablation Experiments. All experiments are run with a shorter training schedule of 10 epochs, causing the results here to be different from the main results. Rows marked in \colorbox{Gray}{gray} indicate options that are used in the main results. \textit{MG} denotes Mask Grounding.}
\label{table3}
\end{table*}

\vspace{-0.1cm}
\subsubsection{Ablating Different Aspects of Mask Grounding}  

\noindent \textbf{Effect on RIS Performance.} In Tab.~\ref{table3}(a), we show that both masked language modeling (MLM) and masked vision-language modeling (MaskedVLM) fail to deliver meaningful performance gains. In contrast, our proposed Mask Grounding improves over MLM by encouraging our model to learn fine-grained visual grounding in language features through usage of additional visual and segmentation information. When added to LAVT, Mask Grounding yields a significant performance gains of 1.44 points on RefCOCO and 1.28 points on RefCOCO+.
\setlength{\intextsep}{1pt plus 1pt minus 1pt}
\begin{figure}[h]
\vskip +0.05 in
\centering
\includegraphics[width=1.0\linewidth]{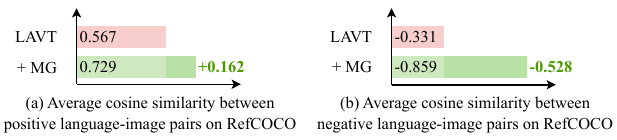}
\vskip -0.1 in
\caption{Mask Grounding Improves Language-Image Alignment.}
\label{figure5}
\vskip +0.1 in
\end{figure}


\noindent \textbf{Effect on Language-Image Alignment.} Next, we check if Mask Grounding can help to improve language-image alignment in RIS models. Effective alignment is indicated by high feature similarity for matching language-image pairs and low similarity for non-matching pairs. To verify this property, we compare the average normalized cosine similarity for all language-image pairs in the RefCOCO validation dataset before and after Mask Grounding is added to LAVT. Since language and image features have different dimensions, we first train a linear layer with contrastive loss to project language features to the same dimension as image features before computing this metric. This method is similar to linear-probing widely used in self-supervised learning \cite{moco, simsiam}. All other weights are frozen in the process. As illustrated in Fig.~\ref{figure5}, Mask Grounding can indeed significantly improve language-image alignment in existing RIS models.

\noindent \textbf{Mask Encoder Design.} In Tab.~\ref{table3}(b), we compare passing two different types of mask input to the mask encoder in Mask Grounding. As shown, using center coordinates of masked region gives slightly better performance. 

\noindent \textbf{Mask Token Predictor Design.} In Tab.~\ref{table3}(c), we evaluate the sensitivity of Mask Grounding's masked token predictor to different MLP layers. We observe that a sufficiently deep masked token predictor is important for good performance. As shown, performance is the best when 8 layers are used. When more layers are added, performance slightly drops, as the model starts to overfit to the auxiliary task.

\noindent \textbf{Universality of Mask Grounding.} In Tab.~\ref{table3}(d), we show that Mask Grounding is also compatible with other representative RIS methods. As shown, when Mask Grounding is added, on RefCOCO and RefCOCO+, we can obtain a performance gain of 0.89 and 1.29 points for CRIS and a performance gain of of 1.10 and 1.52 points for ReLA.

\vspace{-0.2cm}
\subsubsection{Ablating Other Components of Our Method}

\noindent \textbf{Effectiveness of CAM.} \underline{C}ross-modal \underline{A}lignment \underline{M}odule (CAM) improves language-image alignment by injecting global contextual prior into image features. As shown in Tab.~\ref{table3}(e), when CAM is used, we can improve RefCOCO's and RefCOCO+'s oIoU by 0.98 and 1.06 points respectively. Additionally, Tab.~\ref{table3}(e) also shows that using more pyramid scales is helpful in boosting performance.

\noindent \textbf{Effectiveness of CAL.} \underline{C}ross-modal \underline{A}lignment \underline{L}oss (CAL) provides additional pixel-to-pixel ($\mathcal{L}_\text{P2P}$) and pixel-to-text ($\mathcal{L}_\text{P2T}$) alignment supervision to further reduce language-image modality gap. As shown in Tab.~\ref{table3}(f), both $\mathcal{L}_\text{P2P}$ and $\mathcal{L}_\text{P2T}$ alone can bring noticeable oIoU improvement on RefCOCO and RefCOCO+. When the both are added together, we can surpass the baseline by 1.36 points on RefCOCO and 1.63 points on RefCOCO+.

\begin{figure}[h]
\centering
\includegraphics[width=0.9\linewidth]{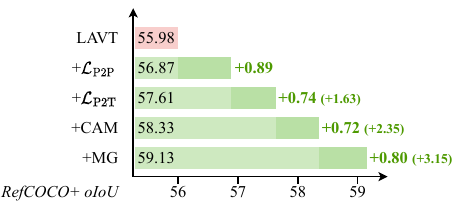}
\vskip -0.05 in
\caption{Compatibility of MagNet components.}
\label{figure6}
\end{figure}

\noindent \textbf{Compatibility of all MagNet components.} As shown in Fig.~\ref{figure6}, all components of MagNet are highly compatible as they progressively improves the LAVT's performance when added incrementally. When all components are added, we can improve LAVT by 3.15 points on RefCOCO+.

\section{Conclusion}
In this paper, we present Mask Grounding, an novel method designed to enhance RIS by teaching our model to predict randomly masked textual tokens based on their surrounding textual, visual and segmentation information. This task requires our model to learn fine-grained visual-textual object correspondence, thus learning visual grounding in the process. When plugged into existing RIS algorithms, Mask Grounding can improve their performance consistently. To holistically address the modality gap, we also design a cross-modal alignment loss and an accompanying alignment module. When all these techniques are used together, our newly proposed MagNet achieves SOTA performance in all RIS benchmarks. We believe that Mask Grounding can also be used in other multi-modal dense prediction tasks and will explore that in future work.

\noindent \textbf{Acknowledgements.} We thank Wan Ding and Kai Ding for their kind support in this project. This work is supported in part by the National Key R\&D Program of China under Grant 2021ZD0140407, the National Natural Science Foundation of China under Grants 62321005 and 62276150, and the THU-Bosch JCML.

{
    \small
    \bibliographystyle{ieeenat_fullname}
    \bibliography{main}
}

\end{document}